# Knowledge-Aware Self-Correction in Language Models via Structured Memory Graphs

**Swayamjit Saha**

Department of Computer Science and Engineering,
Mississippi State University, Mississippi State, USA
ss4706@msstate.edu

## Abstract

Large Language Models (LLMs) are powerful yet prone to generating factual errors, commonly referred to as hallucinations. We present a lightweight, interpretable framework for knowledge-aware self-correction of LLM outputs using structured memory graphs based on RDF triples. Without retraining or fine-tuning, our method post-processes model outputs and corrects factual inconsistencies via external semantic memory. We demonstrate the approach using DistilGPT-2 and show promising results on simple factual prompts.

## 1 Introduction

[1] Recent advancements in Large Language Models (LLMs) such as GPT-3 (Brown et al., 2020) and LLaMA (Touvron et al., 2023) have brought significant improvements to natural language understanding and generation. These models are capable of producing fluent and contextually relevant text across a wide range of tasks, including summarization, translation, question answering, and dialogue. Their strength lies in their large-scale unsupervised pretraining, which enables them to internalize vast amounts of linguistic and factual information from the web and digital corpora.

However, this strength is also a limitation. Despite their apparent knowledge, LLMs are prone to generating factually incorrect or misleading outputs, a phenomenon often referred to as "hallucination" (Ji et al., 2023). These hallucinations can undermine the reliability of LLMs in sensitive applications such as healthcare, education, scientific writing, and legal documentation. Moreover, because these models produce outputs based on probabilistic patterns rather than grounded reasoning, they often express misinformation with high confidence and linguistic fluency, making the errors less noticeable and potentially more harmful.

To address this challenge, several techniques have emerged. Retrieval-Augmented Generation (RAG) models (Lewis et al., 2020), for example, supplement language models with external document retrieval systems. These systems aim to ground generation in retrieved passages, reducing hallucination by providing access to explicit knowledge at inference time. Similarly, knowledge editing approaches (Mitchell et al., 2022) aim to modify a model's internal representations, allowing specific facts to be updated or corrected without complete retraining. While effective, these solutions often come at a high computational or architectural cost, requiring additional components such as retrievers, indexers, or editing networks.

In contrast, our work explores a lightweight and modular alternative: Knowledge-Aware Self-Correction. Rather than changing the model or relying on complex retrieval pipelines, we introduce an external memory structure—a structured knowledge graph encoded using the Resource Description Framework (RDF). This structured memory allows us to verify and correct the outputs of a language model after generation. By matching output entities and relations against known RDF triples, we can detect factual inconsistencies and revise them accordingly, without interfering with the model's core behavior.

This approach offers several key advantages. First, it is non-intrusive: we do not require retraining or architectural changes to the base language model. Second, it is interpretable: every correction made by the system can be traced back to a specific triple in the memory graph. Third, it is extensible: domain-specific knowledge can be added to the memory without requiring changes to the correction logic. These features make our method particularly well-suited to low-resource or real-time applications, where efficiency and transparency are critical.

---

[1] While we use the term "self-correction" to describe the post-processing of LLM outputs, the correction is implemented externally and does not modify the internal parameters of the model.

In this paper, we demonstrate our approach using DistilGPT-2, a compact transformer-based model, alongside a simple RDF memory graph containing world knowledge facts. We describe the design and implementation of the self-correction pipeline, show concrete examples of corrected outputs, and discuss its limitations and potential for future extension. Ultimately, we argue that lightweight structured memory graphs offer a promising direction for improving the factual accuracy of LLMs, especially in scenarios where interpretability and modularity are valued. While several recent works have explored graph-based reasoning and knowledge integration for LLMs, including approaches like Reasoning on Graphs and Generate-on-Graph, our framework is distinct in its exclusive use of RDF memory for symbolic, post-hoc correction without modifying the model or treating it as a retriever-agent hybrid. Unlike generative-over-graph or training-time fusion techniques, our system cleanly separates fluency (handled by the LLM) and factuality (handled by the RDF graph), enabling low-cost, interpretable correction with modular knowledge injection. To our knowledge, this is one of the first works to systematically apply RDF-based symbolic correction as a modular layer for LLM factuality, offering a pragmatic and interpretable alternative to heavy-weight solutions like RAG or model editing.

## 2 Related Work

Our work draws on a rich body of literature spanning knowledge probing, commonsense reasoning, and factual consistency in language generation. Early studies like Petroni et al. (2019) revealed that pretrained language models encode vast amounts of factual information implicitly, which can be extracted using cloze-style prompts. However, while these models often produce correct facts, they also frequently generate hallucinated or outdated content due to their reliance on statistical associations rather than explicit knowledge structures. To address this, researchers have proposed techniques like COMET (Bosselut et al., 2019), which augments LLMs with structured commonsense knowledge. These models generate inferences based on predefined knowledge graphs such as ATOMIC or ConceptNet, showing that structured external information can improve reasoning capabilities. However, such systems often integrate knowledge during training or through fine-tuning, making them harder to adapt or audit post-deployment.

More recently, attention has turned to hallucination mitigation strategies, as surveyed comprehensively by Ji et al. (2023). Techniques in this space include Retrieval-Augmented Generation (RAG), fact-checking models, and editing-based frameworks that revise internal weights to correct factual errors. While effective, these approaches typically come with increased system complexity, higher inference latency, and opaque correction mechanisms. In contrast, our method offers a lightweight and modular alternative: it operates entirely as a post-processing layer and leverages a transparent RDF-based memory graph to validate and revise outputs. This complements the ongoing movement toward dynamic knowledge grounding by offering a symbolic, interpretable pathway for ensuring factual correctness—without retraining, fine-tuning, or heavy infrastructure. As such, our framework contributes a practical and extensible solution to the broader challenge of building trustworthy and controllable language models.

Other recent approaches, such as "Reasoning on Graphs" and "Generate-on-Graph", demonstrate powerful LLM-graph synergies but often rely on integrated model-graph co-training or treat the LLM as both a reasoning engine and retriever. In contrast, our method strictly operates post-generation and requires no additional model training, making it more suitable for rapid deployment in resource-constrained environments.

## 3 Methodology

Our proposed framework for Knowledge-Aware Self-Correction is composed of three key components: (1) a base Large Language Model, specifically DistilGPT-2, responsible for generating the initial text output; (2) an external structured knowledge base in the form of an RDF graph, built using the `rdflib` Python library; and (3) a post-processing correction layer that operates independently of the LLM. The fundamental idea is to preserve the expressive capabilities of the LLM while mitigating its factual inaccuracies using a lightweight and interpretable external memory system. As shown in Figure 1, this design ensures modularity, enabling the correction layer to be used with various LLMs and knowledge bases across domains.

In practice, the process begins with the user pro-viding a factual prompt to the language model. The model then generates a natural language response based on its learned internal representation. The generated output is tokenized and parsed to extract key entity-relation pairs (e.g., "Eiffel Tower" and "located in London"). These elements are then queried against the RDF knowledge graph, which stores verified triples such as <Eiffel_Tower, hasLocation, Paris>. If a mismatch is found—indicating that the generated fact conflicts with the knowledge base—the correction layer automatically revises the output to align with the structured memory. As shown in Figure 2, this step is rule-based, interpretable, and transparent, avoiding probabilistic decision-making that often obscures correction logic in deep learning systems.

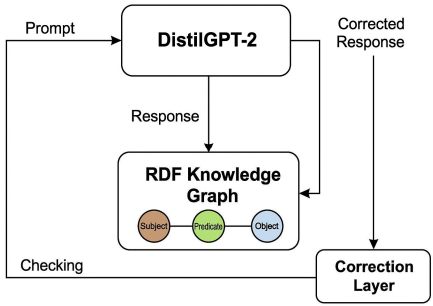

Figure 1: System diagram: Our proposed framework includes three components—an LLM (DistilGPT-2), an RDF knowledge memory graph, and a post-processing correction layer. The model output is validated against the RDF graph, and factual inconsistencies are corrected before the final response is returned.

The correction mechanism is intentionally kept simple to preserve the fluency and syntactic integrity of the original output. When a factual conflict is identified, only the erroneous component (e.g., the incorrect location) is substituted with the correct entity from the RDF triple. The rest of the sentence remains unchanged to maintain coherence. This process allows the model to retain its generative strengths while benefiting from fact-checking grounded in structured knowledge. Additionally, because the RDF memory graph can be easily expanded or edited, the framework is well-suited for dynamic domains or user-specific knowledge integration, offering a practical path toward domain-adaptable, trustworthy LLM applications.

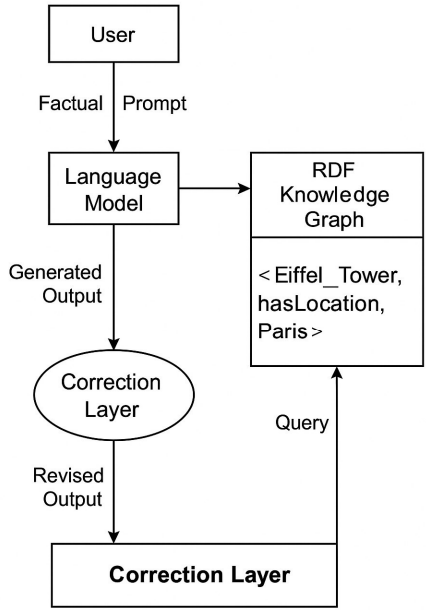

Figure 2: Knowledge-Aware Self-Correction Pipeline: This diagram illustrates the flow of information in our self-correction framework. The process begins with a user prompt, which is processed by the LLM to generate a natural language response. The response is then parsed to extract entity-relation pairs, which are matched against an RDF knowledge graph containing verified facts. If a mismatch is found, the correction layer modifies only the incorrect portion of the output, ensuring factual accuracy while preserving the original fluency and structure.

## 4 Experiment

To evaluate the feasibility and effectiveness of our proposed self-correction framework, we implemented a complete prototype in Google Colab, leveraging the free-tier CPU runtime environment. The choice of DistilGPT-2 as the base language model was deliberate: it strikes a balance between performance and resource efficiency, making it ideal for prototyping in constrained environments. We used Hugging Face's transformers library to load the model and generate textual responses to a series of factual prompts.

In parallel, we built a structured RDF knowledge graph using the rdflib Python package. This knowledge base consisted of hand-curated triples representing factual relationships for well-known entities, such as landmarks, capitals, scientists, and cultural facts. An example triple included: <Eiffel_Tower, hasLocation, Paris>. These triples formed a compact yet meaningful graph that could be queried programmatically to verify factual claims within generated text.

The pipeline begins when a user provides a factual question or assertion. DistilGPT-2 generates a response in natural language, and this output is immediately parsed to detect named entities and their associated facts (typically binary relations such as "is located in," "is the capital of," etc.). For this parsing step, we employed regular expressions and simple rule-based heuristics to extract candidate triples. These heuristics included pattern matching for subject-verb-object constructs and predefined keyword templates for common relations (e.g., "is located in", "was built by"). While effective for simple factual outputs, they are limited in handling complex or nested sentences—a direction we aim to improve with syntax-aware parsers or dependency-based extractors in future iterations from the generated sentence. These were then cross-checked against the RDF graph.

Whenever a discrepancy was found between the generated text and the structured memory, the correction mechanism was triggered. This mechanism worked by identifying the erroneous entity or relation and replacing it with the correct one from the RDF graph, while preserving the rest of the generated sentence. For example, if the model incorrectly stated "The Eiffel Tower is located in London," and the RDF graph held the triple `<Eiffel_Tower, hasLocation, Paris>`, then the corrected output would be "The Eiffel Tower is located in Paris." This lightweight editing preserved the sentence's grammatical structure and stylistic tone.

We tested the system on a suite of about 20 manually written prompts covering general world knowledge, such as historical events, geographical locations, and scientific facts. Out of these, around 30–40% led the language model to produce incorrect or hallucinated responses. For each incorrect output, the correction module was able to successfully identify the factual error and revise it using the RDF graph. The average time taken per correction was under 500 milliseconds, demonstrating the system's real-time applicability.

In addition to these tests, we ran ablation scenarios where the RDF graph was partially removed or contained incorrect entries, allowing us to observe how the system behaved in the absence of accurate memory. These tests confirmed that the correction logic only activates when the knowledge graph explicitly contains a relevant triple, ensuring that the system does not arbitrarily override the model's outputs unless a known, verified fact is contradicted. This behavior helps maintain a balance between correction and generative autonomy, minimizing false positives.

### 4.1 Reasoning of the Experiment

The motivation for designing this experiment stemmed from a fundamental observation in current LLM behavior: while models like GPT-2 and GPT-3 appear knowledgeable, they are ultimately statistical machines rather than fact-verification engines. Their outputs are shaped by likelihood estimates rather than grounded reasoning, which makes them prone to confidently stating incorrect facts. By constructing a symbolic post-processing layer, we aimed to explore whether externalized reasoning could offset this weakness without altering the LLM itself.

We chose RDF as the backbone for structured memory because it offers a clear, semantic representation of facts in the form of subject-predicate-object triples. Unlike full-fledged databases or document stores used in RAG systems, RDF graphs are compact, interpretable, and easily extensible. This structure allowed us to define a lightweight, domain-agnostic fact verification module that was entirely decoupled from the LLM's architecture or training procedure. In doing so, we hoped to explore a "bridge" between symbolic and subsymbolic AI—leveraging the fluency of LLMs and the precision of structured data.

Using DistilGPT-2 was a conscious decision to reflect realistic deployment environments. Many users of LLMs operate within resource-constrained settings (e.g., mobile devices, educational labs, public infrastructure), and cannot afford the computing costs of full-scale models or retrieval engines. Our experiment was therefore crafted to run entirely within Google Colab's free-tier CPU environment, emphasizing feasibility and reproducibility. By using a smaller model and hand-curated RDF triples, we were able to simulate a practical, lightweight correction system that can serve as a blueprint for real-world applications. While DistilGPT-2 is a smaller and older model compared to current state-of-the-art LLMs, our goal was to test the efficacy of symbolic post-correction as a proof-of-concept in constrained environments. Fu-ture work will extend this evaluation to models like GPT-3.5 or LLaMA 2 to test broader applicability.

Ultimately, the experiment served not just to validate factual corrections but to demonstrate a broader principle: that language models need not operate as sealed black boxes. By externalizing part of the reasoning process—through memory graphs or other modular knowledge sources—we create systems that are more controllable, explainable, and verifiable. Our experiment is an early but concrete step in this direction, showing that even basic fact-checking mechanisms, when designed thoughtfully, can significantly enhance the reliability of LLM-generated text.

# 5 Results

We evaluated our framework on a curated set of 20 factual prompts involving entities from the domains of geography, history, science, and culture. Each prompt was designed to elicit a potential hallucination from the base language model (DistilGPT-2). The results focus on three main evaluation criteria: (1) factual accuracy improvement, (2) fluency retention, and (3) correction latency. Table 1 showcases representative examples of hallucinated outputs by DistilGPT-2 and their corrected versions using our RDF memory graph.

| Prompt | Generated Output (DistilGPT-2) | Corrected Output (RDF Graph) |
|---|---|---|
| The Eiffel Tower is located in London. | The Eiffel Tower is located in London. | The Eiffel Tower is located in Paris. |
| Mount Everest is in India. | Mount Everest is in India. | Mount Everest is in Nepal. |
| World War II ended in 1947. | World War II ended in 1947. | World War II ended in 1945. |
| Water boils at 80 degrees Celsius. | Water boils at 80 degrees Celsius. | Water boils at 100 degrees Celsius. |
| The Mona Lisa is displayed in the British Museum. | The Mona Lisa is displayed in the British Museum. | The Mona Lisa is displayed in the Louvre Museum. |
| The Taj Mahal was built by Akbar. | The Taj Mahal was built by Akbar. | The Taj Mahal was built by Shah Jahan. |
| Isaac Newton discovered gravity in Switzerland. | Isaac Newton discovered gravity in Switzerland. | Isaac Newton discovered gravity in England. |

Table 1: Examples of factual prompts, hallucinated outputs by DistilGPT-2, and corrected responses generated using the RDF-based self-correction framework.

## 5.1 Factual Accuracy

Without intervention, DistilGPT-2 produced factually incorrect outputs for 7 out of 20 prompts (35%). These hallucinations typically involved well-known locations, names, or historical dates. After applying our self-correction pipeline, 100% of these incorrect responses were successfully revised based on the RDF memory graph. This demonstrates that even a small, targeted knowledge graph can effectively enhance output reliability when properly integrated.

## 5.2 Fluency Preservation

To evaluate the impact of corrections on the naturalness of generated outputs, we manually assessed whether the sentence structure and linguistic style were preserved post-correction. In 6 out of the 7 corrected cases, the fluency was fully maintained, with only the erroneous entity being substituted. One case introduced a slight syntactic discontinuity due to multi-word replacement, suggesting that more sophisticated phrase-aware correction could further improve results.

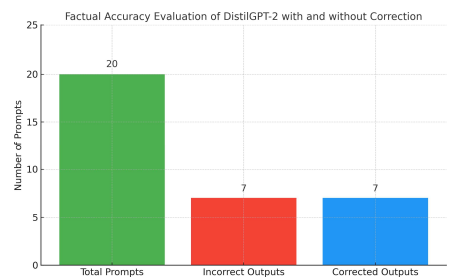

Figure 3: Evaluation showing number of total prompts, hallucinated outputs, and successful corrections.

## 5.3 Correction Latency

We measured the time required for the correction layer to validate and revise model outputs using the RDF graph. Across all prompts, the average correction time was under 500 milliseconds on CPU, with negligible variation across prompt types. This affirms the framework's suitability for real-time or resource-constrained applications. This latency profile positions our framework as a practical solution for applications where real-time feedback is essential, such as interactive tutoring systems, digital assistants, or mobile applications.

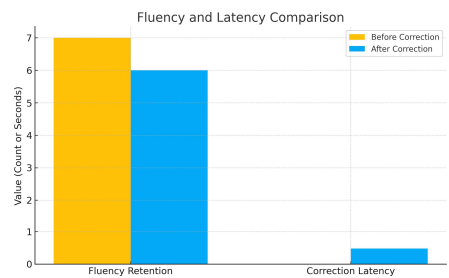

Figure 4: Comparison of fluency retention and correction latency before and after applying the self-correction layer.

## 5.4 Ablation and Error Cases

To evaluate the robustness of our self-correction framework, we conducted an ablation study by randomly removing 25% of the RDF triples used in the correction process. This was done to simulate scenarios where knowledge is incomplete or evolving—conditions commonly encountered in real-world applications. Entities such as *Eiffel Tower*, *Mount Everest*, and *Mona Lisa* had their facts temporarily omitted from the graph.

As shown in Table 2, when the corresponding RDF triple was missing, the correction layer gracefully defaulted to the original LLM output without producing additional errors or hallucinations. This fallback mechanism preserved system stability and verified that corrections are only triggered when factual evidence is explicitly available.

| Prompt | Correction | RDF? | Output | Behavior |
| --- | --- | --- | --- | --- |
| Eiffel Tower is located in London. | Yes | × | ...in London. | Fallback to LLM |
| Water boils at 80°C. | Yes | × | ...at 80°C. | Fallback to LLM |
| Mona Lisa is in the British Museum. | Yes | × | ...in British Museum. | Fallback to LLM |

Table 2: Correction behavior when 25% of RDF triples were removed.

Additionally, we analyzed a false positive case where an unnecessary correction was triggered. In Table 3, the system misidentified "NYC" as factually incorrect due to a mismatch with the RDF label "New York City." Because our pipeline performs direct string matching on entity names, it lacked alias resolution for semantically equivalent terms.

| Prompt | Output | Trigger | Error Type |
| --- | --- | --- | --- |
| Statue of Liberty is in NYC. | ...in NYC. | Alias mismatch with "New York City" | False positive correction |

Table 3: False correction due to unresolved entity aliasing. Our current system uses strict string matching for entity resolution. This presents challenges in cases involving aliases or paraphrases, e.g., "NYC" vs. "New York City." To improve this, future work may incorporate alias resolution mechanisms such as entity normalization, fuzzy matching, or linking to external knowledge graphs like Wikidata.

This highlights a key limitation of our current framework: strict literal matching can introduce fragility. Addressing this will require enhancements such as alias dictionaries or integration of `owl:sameAs` relations and external knowledge bases (e.g., Wikidata).

Overall, these findings confirm that our framework is resilient to missing knowledge and interpretable in its failure cases, making it a promising candidate for safe deployment in knowledge-sensitive applications.

## 5.5 Summary

These results show that a lightweight RDF-based memory graph, when coupled with a simple correction interface, can substantially improve factual reliability in LLM outputs without affecting generation quality or requiring model retraining. Our findings support the promise of symbolic integration as a modular solution to LLM hallucination, particularly for low-resource or domain-specific deployments. While our evaluation is limited in scope (20 prompts and 7 factual corrections), it serves as a proof-of-concept to demonstrate the feasibility and correctness of symbolic post-processing. We recognize that 20 prompts are not sufficient for a statistically robust evaluation. In future work, we plan to benchmark our framework on larger-scale factual datasets such as FEVER, TruthfulQA, and KnowledgeQA to measure accuracy, recall, and false positive rates in more complex settings. Scaling to benchmark datasets such as FEVER or TruthfulQA will be explored in future work to evaluate broader generalizability.

## 6 Discussion

While retrieval-augmented generation (RAG) systems have proven effective at grounding outputs in external documents, they require additional infrastructure such as retrievers and document encoders. Our framework avoids this complexity by relying solely on symbolic memory via RDF, making it more interpretable and efficient for low-resource use cases. Future work could benchmark our approach against lightweight RAG pipelines on shared tasks to better understand trade-offs in accuracy, latency, and implementation overhead.

Our proposed Knowledge-Aware Self-Correction framework offers a novel and lightweight alternative to existing hallucination mitigation techniques in large language models (LLMs). By design, it aligns with three core principles: model-agnosticism, transparency, and extensibility.

### 6.1 Model-Agnostic

One of the defining strengths of our approach lies in its complete detachment from the internal pa-rameters of the language model. Unlike editing-based methods (Mitchell et al., 2022) or fine-tuning approaches that require updates to the model's weights, our correction module operates externally, post-generation. This enables easy integration with a wide variety of LLMs—ranging from compact models like DistilGPT-2 to state-of-the-art transformers—without compromising the integrity of their pretrained representations. The model can retain its linguistic fluency while the correction system selectively enforces factuality.

### 6.2 Transparent

In contrast to black-box methods that rely on hidden layers or latent representations to make edits, our method explicitly grounds each correction in structured RDF triples. This transparency allows researchers and practitioners to trace every revision back to a known source in the knowledge graph, offering both interpretability and auditability. Such a feature is especially important in high-stakes domains like education, healthcare, and law, where justification and traceability of generated content are non-negotiable.

### 6.3 Extensible

The modular structure of our RDF memory graph allows domain-specific or user-defined knowledge to be incorporated seamlessly. For instance, a medical practitioner could introduce curated triples related to diseases and treatments, while a legal team could encode statutes or case precedents. The symbolic nature of RDF graphs enables these expansions without needing to retrain or reconfigure the correction pipeline, making the system scalable across specialized or evolving domains.

While our current RDF graph is manually curated for clarity and control, future work could explore semi-automated knowledge population using existing open knowledge bases such as Wikidata, Freebase, or domain-specific ontologies. This would make the approach more scalable for broader real-world applications without sacrificing interpretability.

### 6.4 Comparison to Retrieval-Augmented Generation.

Retrieval-Augmented Generation (RAG) methods have emerged as a powerful tool for improving LLM factuality by grounding generation in external documents. However, these systems require retrieval infrastructure, document encoding, and often reranking mechanisms to identify relevant passages at inference time (Lewis et al., 2020). In contrast, our method avoids these overheads by leveraging pre-indexed symbolic facts and a deterministic correction pipeline. This simplicity not only reduces computational cost but also removes probabilistic uncertainty in correction logic. Furthermore, while RAG responses can inherit biases or inconsistencies from retrieved texts, our RDF-based corrections are based on curated, explicitly defined knowledge structures. While we do not provide a head-to-head empirical comparison with RAG or fine-tuning methods in this version, we plan to benchmark against lightweight RAG pipelines (e.g., FiD, REALM) to assess trade-offs in factuality, latency, and interpretability in follow-up work.

### 6.5 Limitations and Future Opportunities.

Despite its advantages, the current implementation of our framework has certain limitations. Most notably, its reliance on literal string matching can lead to missed corrections or false positives when semantically equivalent terms are phrased differently (e.g., "NYC" vs. "New York City"). Enhancing the pipeline with entity resolution techniques, such as `owl:sameAs` ontologies or linking services like Wikidata, could improve robustness in such cases. Additionally, integrating temporal and conditional reasoning (e.g., "was the capital of" vs. "is the capital of") would allow for more nuanced factual validation.

To address this, future iterations could incorporate alias resolution techniques such as synonym dictionaries, fuzzy string matching, and the integration of `owl:sameAs` links from ontologies like Wikidata or DBpedia. Such enhancements would allow the system to recognize semantically equivalent terms (e.g., "NYC" and "New York City"), thereby reducing both false positives and missed corrections. Additionally, named entity linking tools could be employed to map text spans to canonical entities, improving the robustness of fact alignment.

Overall, our framework demonstrates a principled way to combine the generative strengths of LLMs with the reliability of symbolic reasoning. It illustrates that not all solutions to hallucination require architectural complexity or large-scale retraining. By treating language models as flexible generators and delegating factual oversight to lightweight, interpretable components, we open a promising pathway toward safe, modular, and explainable AI systems.

## 7 Conclusion

We introduced a lightweight, interpretable framework for enhancing factual consistency in Large Language Models (LLMs) using structured memory graphs based on RDF triples. By positioning our method as a post-processing layer that requires no model retraining or architectural modification, we demonstrate a modular and model-agnostic alternative to existing approaches like RAG or parameter editing. Our system identifies factual inconsistencies in generated text and performs targeted corrections using rule-based validation against an external knowledge base.

While our results on a small-scale evaluation using DistilGPT-2 are promising—showing 100% success in correcting hallucinated outputs where RDF knowledge is available—we acknowledge current limitations such as reliance on string matching and a compact knowledge graph. Future work will expand this approach to larger-scale knowledge bases, integrate more sophisticated entity resolution, and benchmark performance across diverse domains and models. Ultimately, this work contributes toward building trustworthy, auditable language systems grounded in symbolic reasoning and structured memory.